\documentclass[11pt]{article}

\usepackage[preprint]{acl}

\usepackage{times}
\usepackage{latexsym}
\usepackage{booktabs}
\usepackage[T1]{fontenc}
\usepackage[utf8]{inputenc}

\usepackage{microtype}

\usepackage{inconsolata}

\usepackage{graphicx}

\title{Accepted Prefixes Are Not All You Need:\\ A Negative Result on PEFT-Based Block-Diffusion Drafting}

\author{
 \textbf{Abdurrahman Javat},
 \textbf{Allan Kazakov}
\\
\\
 Bahçeşehir University, \\Department of Artificial Intelligence \\ 
 \texttt{
   \{abdurrahman.javat,allan.kazakov\}@bahcesehir.edu.tr
 }
}

\begin{document}
\maketitle
\begin{abstract}
Speculative decoding accelerates autoregressive language model inference by using a cheap drafter to propose multiple future tokens and a target model to verify them. A common design goal is therefore to improve draft quality while reducing auxiliary parameters and systems overhead. We study a negative result for this direction through PEFT-BD, a same-backbone speculative decoding method in which a LoRA-like adapter acts as a block-diffusion drafter for an autoregressive verifier. PEFT-BD is motivated by several attractive properties: it avoids tokenizer mismatch, avoids loading a separate draft model, adds only a small number of trainable parameters, and uses a BD3LM-style denoising objective to propose a block of tokens in parallel.

Despite these advantages, PEFT-BD does not yield a practical speedup in our Qwen3-0.6B experiments. Although the method obtains nontrivial accepted prefixes, profiling shows that each speculative step requires an adapter-enabled full-backbone draft pass followed by an adapter-disabled full-backbone verification pass. Thus, the drafter is parameter-efficient but not compute-efficient. Our results isolate a simple but important condition for successful speculative decoding: the drafter must be substantially cheaper to execute than the verifier. Longer accepted prefixes alone cannot compensate when draft computation remains verifier-scale.
\end{abstract}

\section{Introduction}

Autoregressive language models decode one token at a time, making generation latency a core bottleneck in LLM serving. Speculative decoding reduces this bottleneck by using a drafter to propose future tokens and a target model to verify them, preserving the target model's outputs while reducing the number of sequential target-model steps when proposals are accepted \citep{leviathan2023fast,chen2023accelerating}.

The usual premise is that the drafter is not only accurate, but substantially cheaper than the verifier. Existing methods achieve this asymmetry with smaller draft models, prediction heads, feature-level drafters, or specialized multi-token proposal mechanisms \citep{eagle,fastmtp,dflash,dart}. These designs can introduce extra serving complexity, including additional model weights, draft-target alignment issues, and tokenizer compatibility constraints.

We study PEFT-BD, a same-backbone speculative decoding method that attaches a LoRA-like block-diffusion drafter to the verifier. The method is designed to avoid tokenizer mismatch and separate draft-model overhead while proposing $D=16$ future tokens in parallel. Despite this motivation, our experiments show that PEFT-BD does not yield a practical speedup: the adapter drafter still runs a full-backbone pass before the verifier runs its own full-backbone pass.

Our contribution is a focused negative result for same-backbone speculative decoding. We show that PEFT-BD satisfies several desirable design goals---shared tokenizer, no separate draft model, parameter-efficient adaptation, and block-parallel proposal generation---yet fails to provide speedup because its draft path remains verifier-scale. This isolates drafter--verifier compute asymmetry as a necessary practical condition that accepted-prefix length alone cannot replace.

\section{Related Work}

\paragraph{Speculative decoding.}
Speculative decoding accelerates autoregressive inference by using a drafter to propose future tokens that are then verified by the target model \citep{leviathan2023fast,chen2023accelerating}. Its speedup depends on both draft quality and draft cost. PEFT-BD focuses on the latter: it studies a case where the drafter is parameter-efficient and produces nontrivial accepted prefixes, but is not sufficiently cheaper to execute than the verifier.

\paragraph{Lightweight drafters.}
Recent methods improve speculative decoding by making the draft path cheaper than full target-model decoding. EAGLE predicts future features and uses them to construct draft tokens, reducing drafting cost while maintaining high acceptance \citep{eagle,eagle2,eagle3}. FastMTP trains a lightweight multi-token prediction head with shared weights and self-distillation, making it a practical same-model-style baseline with an available training recipe and serving path \citep{fastmtp}. These methods differ architecturally, but they share the same central requirement: the draft path must be compute-light relative to verification.

\paragraph{Parallel and diffusion-style drafting.}
DFlash and DART are closest to PEFT-BD conceptually because they target parallel or diffusion-inspired drafting. DFlash uses a lightweight block-diffusion drafter to propose multiple future tokens in parallel \citep{dflash}. DART similarly aims to reduce drafting cost by predicting multiple future positions with a lightweight draft module and constructing candidates through n-gram-guided tree search \citep{dart}. These methods point toward the kind of low-cost draft architecture suggested by our negative result. However, they were not fair drop-in baselines for our Qwen3-0.6B setting: released DFlash checkpoints do not include Qwen3-0.6B, and DART's released Qwen-family weights start at larger base models and rely on a custom n-gram/tree-search inference path.

\paragraph{Parameter-efficient adaptation.}
PEFT-BD also relates to parameter-efficient fine-tuning, especially LoRA-style adapters \citep{hu2022lora}. Adapters are attractive for speculative decoding because they can share the verifier's tokenizer, vocabulary, and backbone while adding few trainable parameters. Our result shows that this is not enough: a parameter-efficient drafter can still be compute-expensive if it executes the full target backbone.

\section{PEFT-BD}

PEFT-BD is a same-backbone speculative decoding method. It uses the original autoregressive model as the verifier and attaches a LoRA-like adapter as the drafter \citep{hu2022lora}. The adapter is trained by distillation from the verifier on 50k TULU-3 samples \citep{tulu}, with a block-diffusion objective inspired by BD3LM \citep{bd3lm}. In our implementation, the block size is $D=16$.

At inference time, PEFT-BD alternates between two modes of the same model. With the adapter enabled, the model drafts a block of candidate tokens. With the adapter disabled, the original autoregressive model verifies the proposed block. This avoids tokenizer mismatch and removes the need to load a separate draft model.

Given a context $x_{1:t}$, PEFT-BD proceeds as follows:
\begin{enumerate}
    \item Enable the adapter and append a block of $D$ masked future positions.
    \item Denoise the full block in one draft step, producing a candidate block $\hat{x}_{t+1:t+D}$.
    \item Disable the adapter and run the base autoregressive verifier on the proposed block.
    \item Accept the longest prefix of $\hat{x}_{t+1:t+D}$ that matches the verifier.
    \item Append the accepted prefix and repeat until the generation budget is reached.
\end{enumerate}

The motivation is that the drafter is parameter-efficient, shares the verifier's tokenizer and backbone, and proposes multiple tokens in parallel rather than autoregressively. In principle, this should reduce several common sources of speculative decoding overhead: separate draft-model memory, draft-target tokenizer mismatch, and sequential draft generation.

However, the design also creates the central failure mode studied in this paper. Although the adapter adds few trainable parameters, drafting still executes the full transformer backbone. Thus, each speculative step consists of an adapter-enabled full-backbone draft pass followed by an adapter-disabled full-backbone verification pass. PEFT-BD is therefore parameter-efficient but not necessarily compute-efficient.

A simple cost model makes the issue explicit. Let $A$ be the number of accepted tokens per speculative round, $C_d$ the draft cost, and $C_v$ the verification cost. Speculative decoding is useful only when
\[
    \frac{C_d + C_v}{A}
\]
is smaller than the cost of producing the same tokens with the target model alone. Increasing $A$ helps, but only if $C_d$ is sufficiently small. PEFT-BD tests the case where the accepted prefix is nontrivial, but $C_d$ remains close to verifier-scale computation.

\section{Experiments}

\subsection{Setup}

We evaluate PEFT-BD on Qwen3-0.6B instruct model \citep{qwen3}. The drafter is trained as a LoRA-like adapter on top of the same backbone used by the verifier. Unless otherwise stated, we use block size $D=16$, one denoising step, bf16 precision, maximum generation length 256, and exact verification.

We evaluate on held-out TULU-style instruction prompts rendered with the Qwen chat template. The primary serving experiments use native SGLang with tensor parallelism $1$. We also run local HuggingFace diagnostic loops. The SGLang runs provide optimized serving throughput and accepted-prefix summaries, while the local loops provide exact token/block acceptance and greedy-output audit information. We treat these as complementary: SGLang results are used for end-to-end serving performance, while local runs are used for algorithmic diagnostics.

\subsection{Baselines}

FastMTP is the main speculative decoding baseline because it provides a practical training and inference path for our model setting \citep{fastmtp}.

We do not include DFlash or DART as primary baselines
because neither method provides a fair drop-in Qwen3-0.6B baseline for our setting. Released DFlash checkpoints do not include Qwen3-0.6B.
DART releases inference code and Qwen-family draft weights for larger
base models, but not for Qwen3-0.6B; reproducing a Qwen3-0.6B DART
drafter would require reimplementing its training procedure from the
paper and repository rather than using a released checkpoint or runnable
training recipe. Including these methods would therefore change the base
model, serving path, or implementation effort, rather than isolate the
PEFT-BD design choice.

\subsection{Metrics}

We report throughput in output tokens per second and accepted prefix length per speculative round. For local diagnostic runs, we also report token acceptance rate, block acceptance rate, and exact greedy-output audit results when available. For profiling, we measure draft and verification latency separately.

Our main question is not whether PEFT-BD can produce acceptable drafts, but whether those drafts are cheap enough to make speculation worthwhile. Therefore, the key comparison is between accepted prefix length and the actual compute cost of producing each draft. % In particular, we test whether a PEFT drafter with few additional trainable parameters behaves like a lightweight drafter at inference time.

\subsection{Ablations}

We include two small ablations. First, we compare one-step and two-step denoising to test whether additional denoising improves accepted-prefix behavior enough to offset its cost. Second, we compare anchored and unanchored drafting in the local verifier loop to measure whether a simple acceptance-improving heuristic changes the overall conclusion.

These ablations are not intended to exhaustively tune PEFT-BD. They are included to check whether the observed failure is caused by an obvious drafting choice. % As shown in the next section, neither ablation changes the main finding: PEFT-BD's bottleneck is that its draft pass remains a full-backbone computation.

\section{Results}

\subsection{Accepted prefixes do not offset verifier-scale drafting}

\begin{figure}[t]
    \centering
    \includegraphics[width=\linewidth]{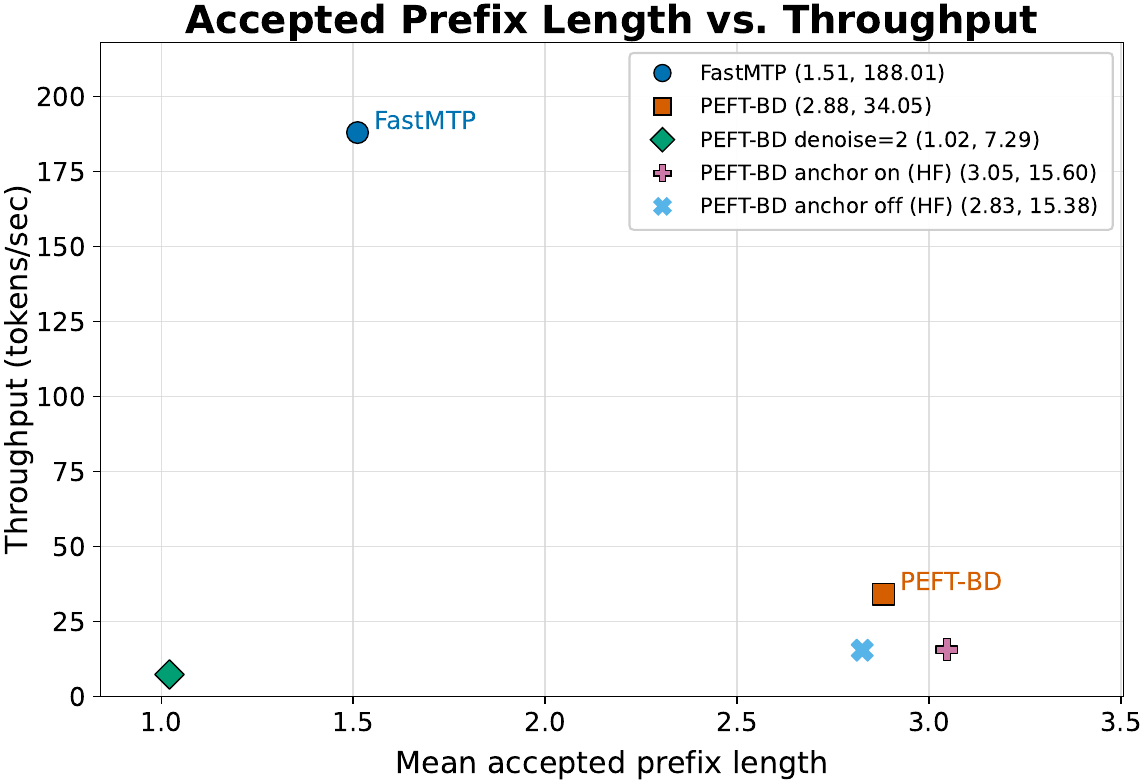}
    \caption{
Mean accepted prefix length versus throughput. All points use native SGLang unless marked with \textsc{HF}. The main native SGLang points report mean throughput over three repeated runs; uncertainty details are in Appendix~\ref{app:uncertainty-oracle}. Optimized SGLang greedy AR is omitted from the plot because our focus is the relation between accepted-prefix length and throughput among speculative methods; greedy-AR diagnostics and small-model context are reported in Appendices~\ref{app:hf_diagnostics}--\ref{app:uncertainty-oracle}.
}
    \label{fig:prefix-throughput}
\end{figure}

Figure~\ref{fig:prefix-throughput} shows the main failure mode of PEFT-BD. In native SGLang evaluation, PEFT-BD obtains a longer mean accepted prefix than FastMTP, but substantially lower throughput. Across three repeated native SGLang runs, PEFT-BD accepts $2.881$ tokens per speculative step on average and reaches $34.05 \pm 0.20$ tokens/s, compared to $1.511$ accepted tokens and $188.01 \pm 9.79$ tokens/s for FastMTP.

This result separates draft quality from draft cost. PEFT-BD is not simply failing to produce plausible proposals: its accepted prefixes are nontrivial, and in this comparison they are longer than FastMTP's. The issue is that each proposal is too expensive to produce. In other words, the relevant metric is not only how many drafted tokens are accepted, but how much computation was spent to obtain those accepted tokens. The repeated runs show that this gap is much larger than runtime variance.

\subsection{The drafter is parameter-efficient but not compute-efficient}

Profiling confirms that PEFT-BD does not provide the compute asymmetry required by speculative decoding. Each speculative step performs two full-backbone passes: one with the adapter enabled for drafting, and one with the adapter disabled for verification. As shown in Table~\ref{tab:profile}, these passes have nearly identical latency.

\begin{table}[h!]
\centering
\small
\begin{tabular}{lcccc}
\toprule
Fwd pass & LoRA & Token width & Calls & $\mu$ latency \\
\midrule
Draft  & Active & 17 & 1769 & 50.285 ms \\
Verify & Inactive  & 17 & 1769 & 50.621 ms \\
\bottomrule
\end{tabular}
\caption{
Forward-level PEFT-BD profile. Drafting costs approximately the same as verification.
}
\label{tab:profile}
\end{table}

A draft-cost oracle gives the same conclusion: we optimistically reduce
only the draft-side full-backbone cost while holding verification cost,
scheduling overhead, and the observed acceptance distribution fixed.
Eliminating the entire draft-side cost would raise PEFT-BD only to
$67.9$ tokens/s, meaning even free drafting would not close the gap to
FastMTP at the observed acceptance rate. The full oracle table appears
in Appendix~\ref{app:uncertainty-oracle}.

\subsection{Ablations do not change the conclusion}

Two small ablations support the same diagnosis. Increasing denoising from one to two steps reduces both throughput and accepted-prefix length, from $34.05$ to $7.29$ tokens/s and from $2.881$ to $1.021$ accepted tokens. In the local HF verifier loop, anchoring slightly improves acceptance, increasing mean accepted prefix from $2.826$ to $3.046$, but throughput remains essentially unchanged at $15.4$--$15.6$ tokens/s. Thus, neither extra denoising nor a simple acceptance-improving heuristic resolves the bottleneck: PEFT-BD remains limited by the expensive full-backbone draft pass.

\section{Discussion}

PEFT-BD isolates a simple failure mode: parameter-efficient drafting is not necessarily compute-efficient drafting. Although the drafter adds few trainable parameters and shares the verifier's tokenizer and backbone, it still executes a full-backbone forward pass. As a result, each speculative step pays for both verifier-scale drafting and verifier-scale verification.

The oracle analysis strengthens this point. Improving acceptance alone is insufficient unless the draft execution path also becomes materially cheaper. In PEFT-BD, accepted-prefix gains are dominated by full-backbone draft computation; under our draft-cost oracle, even aggressive draft-cost reductions would still require much longer accepted prefixes to match stronger baselines.

At larger model scales, absolute throughput may change, but the structural requirement remains: same-backbone speculative methods need real drafter--verifier compute asymmetry. Future designs should therefore optimize the draft execution path itself, for example through layer skipping, early exits, lightweight heads, or state reuse, rather than only reducing trainable parameters or increasing accepted-prefix length.

\section{Conclusion}

We reported a negative result for PEFT-BD, a same-backbone adapter-based block-diffusion drafter for speculative decoding. PEFT-BD has several attractive properties: it shares the verifier's tokenizer and backbone, avoids a separate draft model, adds only a small number of trainable parameters, and produces nontrivial accepted prefixes. However, these properties do not yield practical speedup in our Qwen3-0.6B experiments because the adapter drafter still executes a full-backbone forward pass.

The lesson is simple: accepted tokens are not free. In PEFT-BD, longer accepted prefixes do not compensate for verifier-scale drafting. Same-backbone speculative methods therefore need to make the draft path materially cheaper to execute, not merely more parameter-efficient or more acceptable to the verifier.

\section*{Limitations}

\begin{itemize}
    \item \textbf{Single model scale.}
    Our experiments use Qwen3-0.6B. Larger models, architectures, or hardware may change absolute throughput, but not the structural issue unless the draft path stops scaling with the verifier. Our claim is therefore not that PEFT-BD will have identical slowdowns at every scale, but that same-backbone adapter drafting does not automatically create the drafter--verifier compute asymmetry required for speculative decoding.

    \item \textbf{One PEFT-BD training recipe.}
    We train the adapter on 50k TULU-3 samples with a one-step block-denoising objective. A larger distillation set, different masking schedule, different adapter placement, or different block-diffusion objective could improve acceptance. However, such changes would not remove the central compute issue unless they also make the draft execution path cheaper.

    \item \textbf{Two-step denoising was not trained directly.}
    Our denoise $>1$ ablation should be interpreted cautiously. The adapter was trained for one-step denoising, so poor two-step performance does not imply that multi-step block diffusion is inherently unsuitable for speculative decoding.

    \item \textbf{Limited speculative baselines.}
    FastMTP is our main speculative decoding baseline because it provides a practical training and inference path for our model setting. We do not benchmark DFlash or DART directly because they do not provide
drop-in Qwen3-0.6B checkpoints and serving paths for our setup. In
particular, DART provides inference code and released Qwen-family weights
for larger models, but reproducing a Qwen3-0.6B DART drafter would
require additional training implementation work. Our experiments should
therefore not be read as a broad comparison among speculative decoding
methods.

    \item \textbf{Serving-stack dependence.}
    Throughput depends on the implementation, runtime, batching behavior, kernel support, and availability of optimized execution paths such as CUDA graphs. Our SGLang results are the main serving measurements, while our HuggingFace runs are diagnostic. Different serving stacks could change absolute throughput, although the measured full-backbone draft/verify symmetry is an algorithmic property of PEFT-BD.
    \\

    \item \textbf{Not a negative result for block diffusion or PEFT in general.}
    Our result is specific to a same-backbone adapter drafter that still executes the full verifier backbone. It should not be interpreted as evidence against block diffusion, DFlash-style lightweight drafters, DART-style lightweight modules, EAGLE-style feature drafters, or parameter-efficient adaptation more broadly.
\end{itemize}


\bibliography{custom}

\appendix

\section{Experimental Details}

\subsection{Shared verifier and data contract}

All primary PEFT-BD and FastMTP comparisons use the same frozen target verifier, tokenizer, prompt source, and evaluation protocol. The verifier is Qwen3-0.6B, and prompts are drawn from the TULU-3 SFT mixture and formatted with the Qwen chat template with thinking disabled. Target continuations are generated greedily from the frozen verifier with temperature $0$, maximum prompt length $512$, and maximum continuation length $256$.

Both methods are trained from the same AR-supervised continuation artifact rather than directly from the raw dataset text. The shared split contains 50k training examples, 1k validation examples, and 1k held-out test examples. Verification uses exact greedy top-1 matching and accepts the longest matching prefix.

\subsection{PEFT-BD training}

PEFT-BD trains a LoRA-style adapter on top of the frozen Qwen3-0.6B backbone. The adapter is trained as a BD3LM-style block drafter: future blocks from the shared AR continuation artifact are masked and reconstructed. At inference time, the adapter is enabled only for block-diffusion drafting and disabled for autoregressive verification.

We train PEFT-BD for 10k steps using bf16 precision, SDPA attention, cosine learning-rate decay, learning rate $2\times 10^{-4}$, warmup ratio $0.03$, and effective global batch size $64$. The block size is $D=16$, the mask ratio is sampled from $[0.3,1.0]$, and the prefix loss is the linear-prefix objective with $\gamma=0.9$. The LoRA rank is $16$, LoRA alpha is $32$, and LoRA dropout is $0.0$. LoRA modules are applied to the attention and MLP projection layers. The original AR backbone is frozen; only LoRA parameters and diffusion special-token embeddings are trained.
We apply LoRA to all linear projection modules in the transformer blocks. Thus, the adapter covers both attention projections and MLP projections, while the original backbone weights remain frozen.

Qwen3-0.6B packed with PEFT-BD has 605.9M total parameters and 10.1M trainable parameters, corresponding to 1.666\% trainable parameters. The final checkpoint reaches eval loss 1.3041, eval masked-token accuracy 0.6453, and eval target-token accuracy 0.4413.

\subsection{FastMTP training}

FastMTP is trained from the official FastMTP-compatible path on the same frozen Qwen3-0.6B verifier, same shared AR-supervised continuations, and same held-out test questions. It trains only the MTP layers while freezing the base model and language modeling head.

We train FastMTP for 10k steps using bf16 precision, SDPA attention, cosine learning-rate decay, learning rate $5\times 10^{-5}$, warmup ratio $0.05$, weight decay $0.1$, and effective global batch size $64$. The configuration uses three speculative steps and one next-token prediction layer, with MTP loss weights $[0.51,0.31,0.18]$. Qwen3-0.6B enhanced with FastMTP has 613.9M total parameters and 17.8M trainable parameters, corresponding to 2.905\% trainable parameters.

\begin{table*}[t]
\centering
\normalsize
\begin{tabular}{lccccc}
\toprule
Method & Prompts & Throughput & Mean prefix & Exact audit & Peak VRAM \\
\midrule
Greedy AR & 1000 & 30.729 & 1.000 & -- & 1.344 GiB \\
FastMTP & 1000 & 12.015 & 0.653 & 1.00 & 2.508 GiB \\
PEFT-BD & 1000 & 17.237 & 3.085 & 1.00 & 1.559 GiB \\
\bottomrule
\end{tabular}
\caption{
Local HF diagnostic runs with max generation length 256. 
These results are included for verifier-loop diagnostics and exact greedy-output
audits, not as optimized-serving throughput measurements.
}
\label{tab:hf_diagnostics}
\end{table*}

\subsection{SGLang evaluation}

The primary throughput results use native SGLang serving on the same 1k held-out prompts with maximum generation length $256$, temperature $0$, bf16 dtype, tensor parallel size $1$, and concurrency 1. We report end-to-end serving throughput and accepted-prefix length.

For PEFT-BD, SGLang uses the PEFTBD speculative algorithm with block size $16$, speculative steps $16$, draft token width $17$, Triton LoRA backend, and deterministic adapter switching: the adapter is enabled for block-diffusion drafting and disabled for autoregressive verification. Radix caching is disabled for correctness because PEFT-BD can encounter the same token prefix under different execution modes: adapter-off AR verification and adapter-on BD drafting. The resulting KV states are not interchangeable, so safe prefix reuse would require an adapter-aware and mode-aware cache.

Overlap scheduling and CUDA graphs are disabled in our current PEFT-BD integration because each speculative round uses a nonstandard sequence of forwards: an adapter-on scratch draft pass, restoration of scratch KV/cache state, and then an adapter-off verification pass with a different attention mask. SGLang's optimized overlap and graph-capture paths assume a more stable forward structure, whereas PEFT-BD alternates LoRA state, masks, and temporary cache allocation within a single speculative round.

Although these optimizations would improve absolute throughput, they would not change the algorithmic bottleneck: PEFT-BD still performs a near-full-backbone adapter-enabled draft pass whose latency is comparable to verification.

For FastMTP, SGLang uses the official FastMTP/SGLang path through the EAGLE speculative algorithm, with speculative steps $3$, draft token width $4$, top-$k=1$, and CUDA graphs enabled.

\subsection{Profiling}

The PEFT-BD profiling run uses 20 held-out prompts, maximum generation length $256$, block size $16$, one denoising step, and verification width $17$. The adapter is enabled for drafting and disabled for verification.

Profiling shows that PEFT-BD performs the same number of draft and verification forward calls: 1,769 each. The mean draft forward latency is 50.285 ms, while the mean verification forward latency is 50.621 ms. Thus, the adapter-enabled draft pass costs approximately the same as the adapter-disabled verifier pass. This supports the main claim of the paper: PEFT-BD is parameter-efficient, but its drafter is not compute-efficient.

\subsection{HF Diagnostic Runs}
\label{app:hf_diagnostics}
Table~\ref{tab:hf_diagnostics} reports local HuggingFace diagnostic runs.
These runs expose exact verifier-loop diagnostics and greedy-output audits, but
are not used for optimized-serving throughput claims. 

We do not include optimized SGLang greedy AR in Figure \ref{fig:prefix-throughput} because the
0.6B setting is a small-model regime where highly optimized AR decoding
is faster than the speculative decoders we tested; on our repeated-run setup it reaches $431.36 \pm 4.25$ output tokens/s. We therefore use Figure \ref{fig:prefix-throughput} to isolate
the relation between accepted-prefix length and draft cost among
speculative methods, rather than to claim that speculation is beneficial
at this model scale. The HF greedy AR result in Table \ref{tab:hf_diagnostics} is included only
as a local verifier-loop diagnostic.

\subsection{Repeated-run uncertainty and draft-cost oracle}
\label{app:uncertainty-oracle}

Table~\ref{tab:uncertainty} reports repeated native SGLang runs on the same held-out prompts with deterministic greedy settings. The resulting uncertainty reflects systems/runtime variance rather than data-sampling variance. PEFT-BD remains far below FastMTP across repeated runs despite having a longer accepted prefix. Optimized greedy AR is also included here for context: in this small-model regime, it is faster than both speculative methods, so we omit it from Figure~\ref{fig:prefix-throughput} to keep the main comparison focused on accepted-prefix length versus draft cost among speculative methods.

\begin{table*}[t]
\centering
\normalsize
\begin{tabular}{lrrrr}
\toprule
Method & $n$ & Tok/s & Prefix & Latency (s) \\
\midrule
PEFT-BD & 3 & $34.045 \pm 0.195$ & $2.881 \pm 0.000$ & $7519.504 \pm 43.027$ \\
FastMTP & 3 & $188.011 \pm 9.787$ & $1.511 \pm 0.000$ & $1364.116 \pm 71.935$ \\
Greedy AR & 3 & $431.356 \pm 4.249$ & $1.000 \pm 0.000$ & $593.515 \pm 5.859$ \\
\bottomrule
\end{tabular}
\caption{Repeated native SGLang runs. Values are mean $\pm$ standard deviation over repeated runs on the same held-out prompts.}
\label{tab:uncertainty}
\end{table*}

\begin{table*}[h!]
\centering
\normalsize
\begin{tabular}{rrrr}
\toprule
Draft reduction & Predicted tok/s & Need vs. FastMTP & Need vs. AR \\
\midrule
$0\%$ & $34.045$ & $15.910$ & $36.502$ \\
$25\%$ & $38.890$ & $13.928$ & $31.955$ \\
$50\%$ & $45.343$ & $11.946$ & $27.407$ \\
$75\%$ & $54.364$ & $9.964$ & $22.860$ \\
$90\%$ & $61.732$ & $8.774$ & $20.131$ \\
$100\%$ & $67.865$ & $7.981$ & $18.312$ \\
\bottomrule
\end{tabular}
\caption{Draft-cost oracle for PEFT-BD. Only the draft-side full-backbone cost is reduced; verifier cost, scheduling overhead, and acceptance are held fixed. The last two columns report the accepted-prefix length needed to match FastMTP and optimized greedy AR. The current block size is $D=16$.}
\label{tab:draft-cost-oracle}
\end{table*}

We also compute a draft-cost oracle to separate the effect of draft execution cost from acceptance quality. Let $T_0$ be the measured PEFT-BD iteration time and let $\rho_d$ be the draft-side cost share. For a draft-cost reduction $r \in [0,1]$, we estimate the new iteration time as
$
T(r) = T_0(1 - r\rho_d),
$
while holding verification cost, scheduling overhead, and the observed acceptance distribution fixed. This is an optimistic estimate because it assumes draft-side cost can be reduced without lowering accepted-prefix length or adding new overhead. In our measurements, $T_0 = 0.0846$ s and the draft/verify cost shares are approximately $0.498/0.502$.

The oracle analysis shows that reducing draft cost alone is not sufficient unless accepted-prefix length also increases substantially. Even if the draft-side full-backbone cost were eliminated entirely, PEFT-BD would reach only $67.865$ tokens/s at the observed acceptance rate. Matching FastMTP would require about $8$ accepted tokens per round, while matching optimized greedy AR would require $18.312$ accepted tokens, exceeding the current block size $D=16$.

\end{document}